%% file: main.tex
\documentclass[conference]{IEEEtran}
\IEEEoverridecommandlockouts
\usepackage{cite}
\usepackage{amsmath,amssymb,amsfonts}
\usepackage{graphicx}
\usepackage{algorithm}
\usepackage{algpseudocode}
\usepackage{textcomp}
\usepackage{longtable}
\usepackage{tabularx}
\usepackage{booktabs}
\usepackage{makecell}
\usepackage{xcolor}
\usepackage{booktabs}
\usepackage{multirow}
\def\BibTeX{{\rm B\kern-.05em{\sc i\kern-.025em b}\kern-.08em
    T\kern-.1667em\lower.7ex\hbox{E}\kern-.125emX}}
\begin{document}

\newcommand{\systemname}{CLOUDADV}
\newcommand{\modelname}{Chronos-2}
\newcommand{\cloudprovider}{Azure}

\title{
CLOUDADV: Decision-Aligned Instance Sizing with Zero-Shot Foundation Models under Drift\\
}

\author{\IEEEauthorblockN{1\textsuperscript{st} Jack Bell}
\IEEEauthorblockA{\textit{Department of Computer Science} \\
\textit{University of Pisa}\\
Pisa, Italy \\
jack.bell@di.unipi.it}
\and
\IEEEauthorblockN{2\textsuperscript{nd}  Giacomo Carfi}
\IEEEauthorblockA{\textit{Department of Computer Science} \\
\textit{University of Pisa}\\
Pisa, Italy \\
giacomo.carfi@phd.unipi.it}
\and
\IEEEauthorblockN{3\textsuperscript{rd}  Gerlando Gramaglia}
\IEEEauthorblockA{\textit{Department of Computer Science} \\
\textit{University of Pisa}\\
Pisa, Italy \\
gerlando.gramaglia@phd.unipi.it}
\and
\IEEEauthorblockN{4\textsuperscript{th}  Andrea Simioni}
\IEEEauthorblockA{\textit{Independent Researcher} \\
\textit{Sintra Consulting Limited}\\
Arezzo, Italy \\
a.simioni@sintraconsulting.eu}
\and
\IEEEauthorblockN{5\textsuperscript{th} Daniele Fontani}
\IEEEauthorblockA{\textit{Independent Researcher} \\
\textit{Sintra Consulting Limited}\\
Arezzo, Italy \\
d.fontani@sintraconsulting.eu}
\and
\IEEEauthorblockN{6\textsuperscript{th}  Vincenzo Lomonaco}
\IEEEauthorblockA{\textit{Department of AI, Data and Decision Science} \\
\textit{Luiss University}\\
Rome, Italy \\
vlomonaco@luiss.it}}

\maketitle

\begin{abstract}

Cloud virtual machines are often overprovisioned, creating avoidable cost and operational inefficiency. We present CLOUDADV, an interactive engineer-facing advisory system for cloud instance sizing under workload drift. The system combines zero-shot time-series forecasting with bounded recommendation generation across day-, week-, and month-scale planning horizons. For each query, CLOUDADV constructs a structured decision context from historical utilization, forecast summaries, current VM metadata, candidate instance options, pricing, and explicit sizing heuristics. A higher-capacity LLM is used offline to generate reference recommendations, while a smaller production model is evaluated on the same prompts to assess deployment-time alignment under latency and cost constraints. Evaluation prioritizes downstream recommendation quality using simulated Azure cost savings and ex-post exceedance, with rolling-origin forecast accuracy reported as a secondary diagnostic against classical and supervised baselines. In a case study of seven production VMs, the reference recommendations reduce simulated monthly cost from about \$1,503 to \$708, yielding \$795/month in savings (52.9\%) under conservative heuristic constraints, while the highest observed exceedance rate among downgraded cases is 1.5\%. Although Chronos-2 does not minimize every forecasting metric, it often induces recommendation patterns similar to those of a supervised per-VM baseline. These results suggest that zero-shot foundation models can support decision-aligned provisioning in non-stationary cloud environments while reducing the operational burden of repeated per-tenant retraining, revalidation, and redeployment.

\end{abstract}


\input{paper/1-introduction}
\input{paper/2-related_work}
\input{paper/3-methodology}
\input{paper/4-experimental_setup}
\input{paper/5-results}
\input{paper/7-conclusion}

\section*{Acknowledgements}

This research was partially supported by the FIS2 Grant from Italian Ministry of University and Research (Grant ID: FIS2023-03382). This research was partially supported by the FSE+ 2021-2027 Bando Assegni di Ricerca 2024, funded by the Regione Toscana and co-financed by Sintra Consulting S.r.l. (CUP: I53C23000690004).
\bibliographystyle{IEEEtran}
\bibliography{references}

\newpage
\onecolumn

\end{document}

%% file: paper/1-introduction.tex
\section{Introduction}

Cloud computing enables elastic, on-demand infrastructure, yet VM provisioning in practice is still often conservative and static: resources are sized for anticipated peak demand and then left unchanged for long periods~\cite{singh2016cloud}. This leads to persistent over-provisioning, with cost and sustainability consequences. Large-scale data-centre studies have long shown that servers operate far below peak utilisation for much of their lifetime, while still drawing substantial power when idle or lightly loaded~\cite{barroso2007case}. The problem is becoming more significant as data-centre electricity demand rises, reaching an estimated 415~TWh in 2024 and projected to grow further with increasing AI workloads~\cite{iea2025energyai}. VM right-sizing is therefore not only a cost-optimisation problem, but also a resource-efficiency problem.

\begin{figure}
    \centering
    \includegraphics[width=0.99\linewidth]{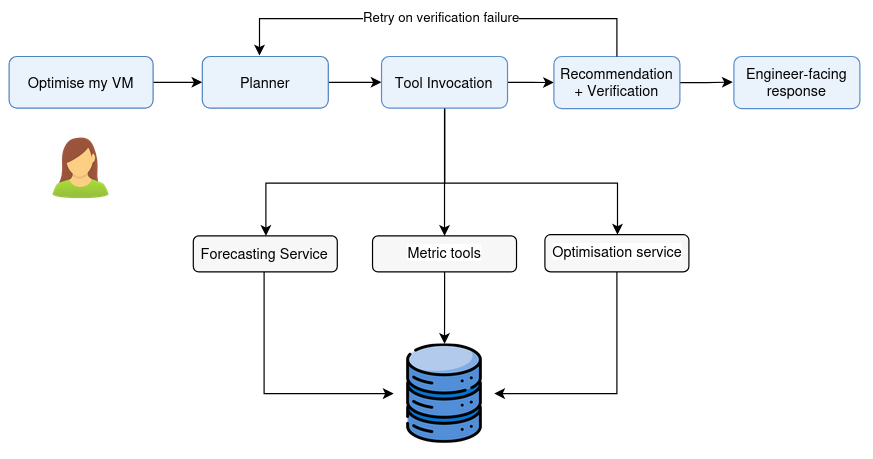}
    \caption{CLOUDADV workflow. A planner routes user requests to forecasting, metrics, and optimisation services. Their outputs are assembled into a bounded decision context for recommendation generation and verification before an engineer-facing response is returned. Verification failures trigger a retry through the tool layer.}
    \label{fig:mobile_interface}
\end{figure}

Addressing this requires an engineer-facing advisory system that supports capacity planning across flexible horizons, with recommendations interpretable enough to assess, validate, and override when necessary~\cite{lipton2018mythos} and without the overhead of maintaining a separate forecasting pipeline per horizon~\cite{taieb2012review}. In the application-server setting studied here, utilisation over time is the primary operational signal for sizing, since direct user-facing latency signals are not available for control.

Recent time-series foundation models offer a promising alternative. Models such as TimesFM~\cite{das2024decoder} and Chronos~\cite{ansari2024chronos} produce zero-shot forecasts at arbitrary horizons from a single pre-trained model, reducing the need for per-horizon or per-tenant training. This is attractive not only for modelling flexibility, but also for continual adaptation under workload drift: because zero-shot models carry no fitted per-VM state, they require no retraining trigger, revalidation pipeline, or redeployment cycle when behaviour shifts. In supervised forecasting systems, changing VM behaviour makes fitted models stale, requiring retraining, revalidation, and redeployment to remain reliable~\cite{losing2018incremental, zhao2025proactive}. 

However, forecasting quality for VM right-sizing cannot be judged only by raw error alone. A forecast may be numerically imperfect yet still preserve the correct provisioning action, while a low-error forecast can still be operationally harmful if it crosses a decision boundary and triggers the wrong scaling recommendation. The relevant criterion is therefore \emph{decision alignment}: whether the forecast is sufficient to preserve the downstream sizing decision \cite{elmachtoub2022smart}. Whether zero-shot time-series foundation models satisfy this requirement in cloud capacity optimisation remains underexplored.

To study this question, we present \systemname{}, an interactive engineer-facing advisory system that combines zero-shot forecasting, structured evidence retrieval, and policy-constrained LLM recommendation generation behind a natural-language interface. We further distinguish between offline reference recommendation generation and deployable online recommendation serving: a higher-capacity model is used to produce structured reference recommendations, while a smaller production model is evaluated on the same bounded prompts to test whether stronger-model recommendation behaviour can be preserved at lower serving cost.

Our contributions are fourfold. First, we formulate cloud instance sizing as a decision-aligned forecasting problem under workload drift. Second, we design \systemname{}, combining zero-shot forecasting, bounded recommendation generation, and engineer-facing interaction for multi-horizon VM sizing. Third, we introduce a deployment-aware reference--production recommendation pipeline for assessing alignment between a stronger offline model and a smaller deployable model. Fourth, we evaluate the system using simulated savings, ex-post exceedance, and forecast accuracy across multiple horizons and model classes.

%% file: paper/2-related_work.tex
\section{Related Work}
\label{sec:related}

\subsection{VM Right-Sizing and Cloud Cost Optimization}

\noindent\textbf{The overprovisioning problem.} VM right-sizing refers to matching a virtual machine's allocated resources to its actual workload demand, and is widely recognised as a primary lever for reducing cloud waste. Industry surveys consistently find that over-provisioning affects the majority of deployments~\cite{cortez2017resource}, with idle resources generating avoidable expenditure that compounds at fleet scale.

\noindent\textbf{Prior work on prediction-informed provisioning.} Cortez et al.~\cite{cortez2017resource}
demonstrated in \emph{Resource Central} that VM workload
behaviours are consistent enough across lifetimes for machine learning to support
prediction-informed scheduling on Azure. Nawrocki and Sus~\cite{nawrocki2022anomaly}
further showed that anomaly detection is a necessary pre-processing step for
long-term capacity planning, since contaminated usage histories lead to poor
provisioning decisions even when the forecasting model is otherwise sound.

\noindent\textbf{Limitations of existing tools.} Commercial tools such as Azure Advisor \cite{microsoft_azure_advisor} and AWS Compute Optimizer~\cite{aws_compute_optimizer_ec2} rely on percentile-based heuristics over fixed lookback windows, providing point-in-time recommendations rather than forward-looking forecasts and no mechanism for engineer feedback. Reactive autoscaling~\cite{lorido2014review} is likewise insufficient, as VM size cannot be changed without explicit reprovisioning. \systemname{} addresses these gaps through probabilistic long-horizon forecasts and an engineer-in-the-loop validation workflow.

\subsection{Forecasting Infrastructure Metrics}

\noindent\textbf{Classical and statistical approaches.} Time series forecasting for infrastructure metrics traditionally relies on statistical models such as \textit{ARIMA}, exponential smoothing, and Prophet~\cite{box2015time,taylor2018forecasting}. These methods are computationally efficient and perform well on stationary series with regular seasonality. However, cloud resource metrics often exhibit regime shifts, bursts, and workload-driven non-stationarity that degrade performance at longer horizons~\cite{nawrocki2022anomaly}.

\noindent\textbf{Deep learning models.} More recent work applies deep sequence models such as LSTMs and temporal convolutional networks, which improve short-horizon forecasting over statistical baselines~\cite{ouhame2021efficient}. TSMixer~\cite{chen2023tsmixer}, a lightweight MLP-mixer architecture, offers competitive accuracy with lower inference cost and serves as one of our supervised baselines.

\noindent\textbf{Challenges at long horizons.} Forecasting accuracy degrades with horizon length, and the practical costs of this degradation differ across applications. For VM right-sizing, the relevant question is not whether the forecast is accurate in an absolute sense, but whether it is accurate enough to preserve the direction of the sizing decision \cite{elmachtoub2022smart}. A forecast that overestimates peak CPU by a modest margin may still recommend the correct SKU, whereas a forecast that misses a sustained workload increase may recommend a downgrade that triggers saturation. This decision-alignment framing, rather than raw error minimisation, motivates our evaluation methodology in Section~\ref{sec:experiments}.

\subsection{Time-Series Foundation Models and Zero-Shot Forecasting}

\noindent\textbf{The emergence of foundation models for time series.} Inspired by the success of large pre-trained models in Natural Language Processing (NLP) and vision, recent work has explored training general-purpose forecasting models on large and diverse corpora of real-world time series \cite{liang2024foundation}. TimesFM~\cite{das2024decoder}, trained by Google on a corpus of hundreds of billions of time points, and Chronos~\cite{ansari2024chronos}, which frames forecasting as language modelling over quantised time series tokens, both demonstrate strong zero-shot performance across a wide range of benchmarks without any task-specific training. This represents a qualitative shift from the classical paradigm, where a separate model must be fitted to each target series or dataset. From an operational standpoint, this shift is particularly valuable under workload drift: a zero-shot model has no fitted parameters to become stale, so it does not require a retraining trigger or revalidation step when VM behaviour changes. Supervised models, by contrast, must detect drift, collect sufficient new data, retrain, and redeploy before their recommendations are again reliable \cite{losing2018incremental, zhao2025proactive} — a cycle that introduces both latency and operational overhead.

\noindent\textbf{Fine-tuning and model scale.} While zero-shot application is the most operationally convenient setting, foundation models can also be fine-tuned on domain-specific data when labelled examples are available~\cite{ansari2024chronos}. We evaluate \modelname{} in its zero-shot configuration as the primary setting, since this reflects realistic deployment constraints where per-tenant training data may be limited or unavailable. 

%% file: paper/3-methodology.tex
\section{Methodology}
\label{sec:problem}

We consider a set of VMs indexed by $i \in \{1,\dots,N\}$. At decision time $t$, each VM $i$ is assigned a current Stock Keeping Unit (SKU) $s_{i,t}$ drawn from a catalogue $\mathcal{S}$. Each SKU $s \in \mathcal{S}$ is characterised by CPU capacity $c^{cpu}(s)$, memory capacity $c^{mem}(s)$, and hourly price $p(s)$. For a planning horizon $H \in \{\text{day}, \text{week}, \text{month}\}$, the goal is to generate a provisioning recommendation for VM $i$ consisting of a proposed future SKU $\tilde{s}_{i,t,H} \in \mathcal{S}$ together with an action label
\[
a_{i,t,H} \in \{\texttt{maintain}, \texttt{downgrade}, \texttt{upgrade}\},
\]
where the action is defined relative to the current SKU $s_{i,t}$.
\begin{figure*}[t]
    \centering
    \includegraphics[width=\textwidth]{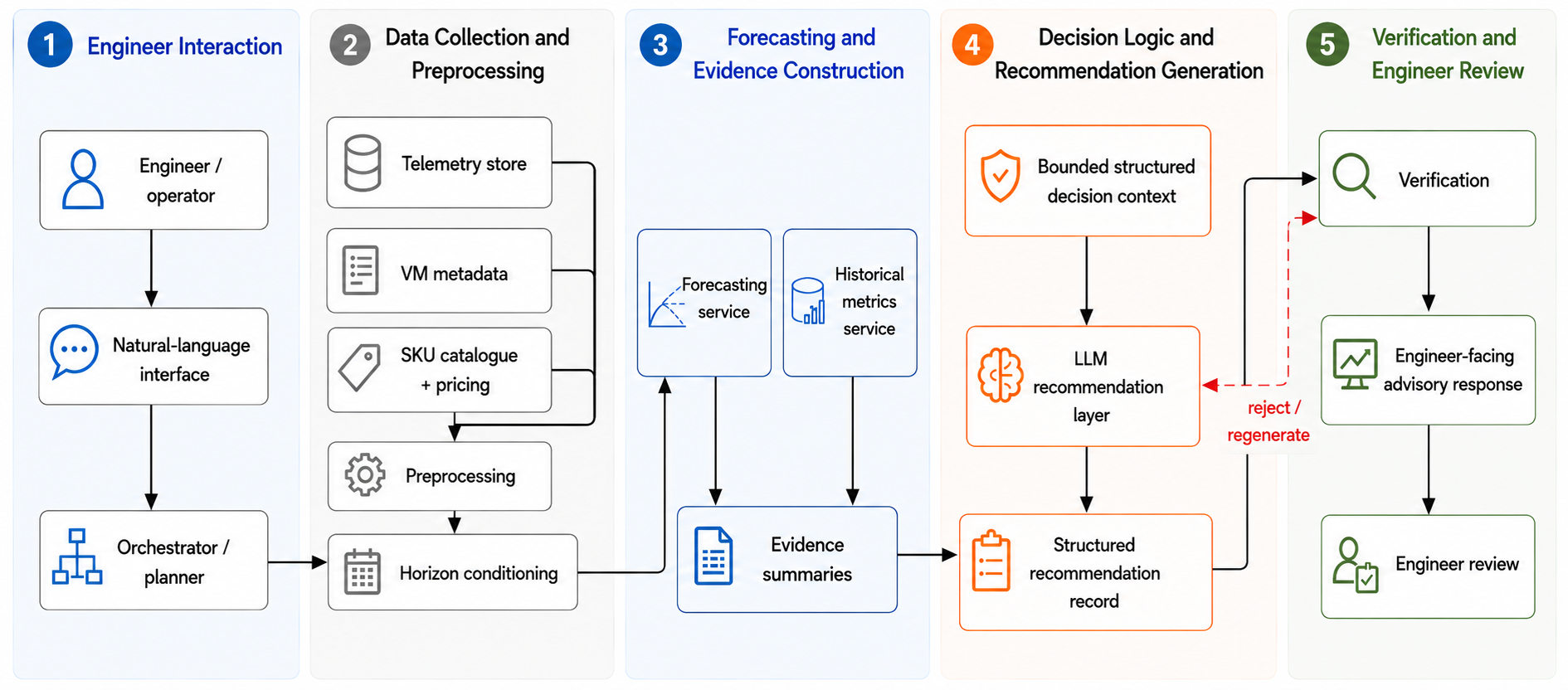}
    \caption{CLOUDADV system architecture. Engineers interact through a natural-language interface while an orchestration layer gathers the information required for The Forecasting service. Historical observations are combined with forecast summaries into a bounded decision context. The recommendation layer generates a structured sizing recommendation, which is verified against the underlying evidence before presentation to the engineer.}
    \label{fig:methodology}
\end{figure*}
\subsection{Operational Objective}

The operational objective is to produce, for each VM and planning horizon,
a safe and cost-effective SKU recommendation grounded in forecast evidence,
current VM characteristics, candidate SKUs, and conservative sizing
heuristics. The problem is therefore decision-oriented: the quantity of
interest is not the forecast itself, but the downstream recommendation
derived within a bounded decision context, which must remain operationally
defensible and consistent with the provided evidence and heuristic
constraints.
\subsection{Telemetry and Preprocessing}

For each VM, we observe percentage CPU utilisation $x^{cpu}_{i,t} \in [0,100]$ and available memory $x^{mem\text{-}avail}_{i,t}$ at 1-minute intervals. Since memory demand is needed in absolute rather than available terms, memory used is derived as
\[
x^{mem\text{-}used}_{i,t} = c^{mem}(s_{i,t}) - x^{mem\text{-}avail}_{i,t},
\]
where $c^{mem}(s_{i,t})$ is the total memory capacity of the VM's current SKU. Before forecasting, each telemetry stream is resampled onto a regular time grid and missing values are imputed by interpolation, yielding aligned CPU and memory time series for downstream analysis.

\subsection{Forecast-to-Decision Pipeline}

Given the processed telemetry for VM $i$ at decision time $t$, CLOUDADV constructs a recommendation for each planning horizon $H \in \mathcal{H}$ through a forecast-to-decision pipeline that couples probabilistic forecasting with bounded recommendation generation. Let $\Delta(H)$ denote the horizon-specific aggregation granularity, $W(H)$ the corresponding historical context window, and $K(H)$ the number of forecast steps associated with horizon $H$. For each resource dimension $r \in \{\mathrm{cpu}, \mathrm{mem}\}$, the telemetry series is first resampled to granularity $\Delta(H)$ over the lookback window $W(H)$, yielding the horizon-conditioned input sequence
\[
x^{(r)}_{i,t,H} = \{x^{(r)}_{i,\tau}\}_{\tau = t-W(H)+1}^{t}.
\]

A forecasting model then produces a probabilistic forecast over the next $K(H)$ steps,
\[
\hat{Y}^{(r)}_{i,t,H} = \{\hat{y}^{(r)}_{i,t+1}, \hat{y}^{(r)}_{i,t+2}, \dots, \hat{y}^{(r)}_{i,t+K(H)}\},
\]
from which the system extracts horizon-relevant summary statistics, including conservative upper-percentile estimates of anticipated future utilization. These forecast summaries are paired with corresponding historical summaries computed over the same resource dimensions so that projected demand can be interpreted relative to recent observed behaviour rather than in isolation.

The forecast outputs are not mapped directly to a final action by a closed-form rule. Instead, the system assembles a \emph{structured decision context}
\[
c_{i,t,H} = \Big(
s_{i,t},
\phi^{\mathrm{hist}}_{i,t,H},
\phi^{\mathrm{fcst}}_{i,t,H},
\mathcal{V}_{i,t},
\mathcal{P}_{i,t},
\Gamma_H
\Big),
\]
where $s_{i,t}$ denotes the current VM configuration, $\phi^{\mathrm{hist}}_{i,t,H}$ and $\phi^{\mathrm{fcst}}_{i,t,H}$ denote historical and forecast-derived utilization summaries, $\mathcal{V}_{i,t}$ denotes the candidate SKU set and associated capacity metadata, $\mathcal{P}_{i,t}$ denotes the relevant pricing information, and $\Gamma_H$ denotes the explicit sizing heuristic associated with horizon $H$. This heuristic encodes the conservative operational constraints used by the system, including headroom requirements, locality restrictions over feasible instance changes, and the preference for cost-efficient recommendations within the bounded candidate space.

The structured decision context is then passed to the recommendation layer, which generates a user-facing recommendation record consisting of an action (e.g., downgrade, maintain, or upgrade), a proposed target configuration, and a rationale grounded in the supplied quantitative evidence. In this formulation, the forecast is an intermediate signal rather than the final object of interest. The primary output of the pipeline is the induced provisioning recommendation supported by the forecast, the historical evidence, and the explicit operational heuristic. This design reflects the fact that, in practice, recommendation quality is determined not only by predictive accuracy, but by whether the resulting action is safe, cost-effective, and defensible under engineer review.

\subsection{Sizing Policy}

For each horizon $H$, the forecasting model produces a probabilistic forecast from which we retain the 95th percentile as a conservative estimate of future demand. To reduce the risk of recommendations being driven by transient forecast underestimation, we define an effective 95th-percentile demand statistic $P_{95,\text{eff}}$ as the maximum of the historical and forecasted 95th percentiles. This yields required CPU and memory capacities
\begin{equation}
    req^{cpu} = \frac{P^{cpu}_{95,\text{eff}}/100}{\tau^{cpu}} \cdot c^{cpu}(s_{i,t}),
    \qquad
    req^{mem} = \frac{P^{mem}_{95,\text{eff}}}{\tau^{mem}},
    \label{eq:req}
\end{equation}
where $\tau^{cpu}$ and $\tau^{mem}$ are configurable headroom parameters controlling the fraction of nominal capacity that may be safely utilised. In all experiments, we set $\tau_{\text{cpu}} = 0.85$, $\tau_{\text{mem}} = 0.80$, and $\delta = 2.0$, so recommended candidates are restricted to SKUs whose CPU and memory capacities lie within a factor of two of the current allocation in both dimensions.

To avoid disruptive resizing actions, candidate SKUs are restricted to a neighbourhood of the current allocation. Specifically, only SKUs whose CPU and memory capacities lie within a factor $\delta$ of the current SKU in both dimensions are considered. Among these candidates, the system selects the cheapest SKU satisfying both
\[
c^{cpu}(s) \ge req^{cpu}
\qquad \text{and} \qquad
c^{mem}(s) \ge req^{mem}.
\]

The selected SKU $\tilde{s}_{i,t,H}$ and the associated feasibility and cost calculations define a bounded recommendation context. In clear-cut cases, this context strongly implies a maintain, downgrade, or upgrade action. In CLOUDADV, however, the final user-facing recommendation is generated by the recommendation layer from this bounded context rather than assigned by a separate closed-form rule. The purpose of the sizing policy is therefore to constrain the feasible decision region and expose the quantitative evidence underlying the recommendation. As a result, in the majority of cases the recommendation action is already strongly implied by the bounded context before it reaches the recommendation layer. The LLM's primary contribution is therefore generating an engineer-readable rationale grounded in the supplied evidence, and resolving borderline cases where the quantitative signal alone does not unambiguously determine the target SKU. This design is intentional: in a safety-critical advisory setting, constraining the decision space before language generation reduces the risk of unsupported recommendations.

\subsection{System Architecture and Recommendation Serving}
\label{sec:system}

CLOUDADV is implemented as an interactive engineer-facing advisory system rather than a batch recommendation pipeline. Engineers interact with the system through a natural-language interface, while an orchestration layer determines which backend capabilities are required for a given query, such as telemetry retrieval, forecasting, recommendation generation, or evidence lookup \cite{yao2022react}. For each VM--horizon request, the system first constructs the structured decision context defined above from historical summaries, forecast summaries, current VM metadata, candidate SKU options, pricing information, and the explicit sizing heuristic.

The recommendation layer is evaluated under two zero-shot model configurations using the same structured prompt regime. Claude Opus 4.6 is used offline as a reference model, generating structured recommendation records from the bounded decision context, while Qwen3.5-35B \cite{qwen35blog} is used online under deployment latency and cost constraints. The production model is not trained, fine-tuned, or distilled to imitate the reference model. Instead, the reference model serves as a stronger benchmark for recommendation quality, and the resulting \emph{reference agreement} measures how closely a deployable model matches stronger-model recommendation behaviour under identical evidence and heuristic constraints.

Because the recommendation layer is user-facing, generated outputs are verified against the underlying structured decision context before presentation \cite{yao2022react}. Identifiers, capacities, prices, and other numeric quantities appearing in the recommendation record must match the retrieved evidence, and outputs that fail these consistency checks are rejected or regenerated. CLOUDADV is designed as an advisory workflow rather than an autonomous controller, so engineers inspect recommendations together with the supporting evidence and may accept, defer, or override them. In evaluation, this is complemented by a stratified engineer audit of the reference recommendation layer to assess both faithfulness to the supplied evidence and operational acceptability in practice.

Algorithm~\ref{alg:cloudadv} summarises the end-to-end recommendation process implemented by \systemname{}. 
For each VM–horizon request, the recommendation layer receives a structured prompt containing: the current SKU identifier and its CPU/memory/price metadata; historical and forecast utilization summaries at the 50th, 95th, and effective 95th percentiles; the filtered candidate SKU list with capacities and prices; and the explicit sizing heuristic for the relevant horizon. The model is instructed to return a structured record with three fields: action, recommended SKU identifier, and a brief rationale referencing the supplied statistics. Outputs are verified against the structured context before presentation, and any identifier or numeric mismatch triggers regeneration.

\begin{algorithm}[t]
\caption{End-to-end recommendation generation}
\label{alg:cloudadv}
\begin{algorithmic}[1]
\Require VM $i$, current SKU $s_{i,t}$, horizon $H$
\Ensure Structured recommendation record $r_{i,t,H}$
\State Load telemetry for context window $W(H)$
\State Resample to a regular time grid and impute missing values
\State Resample to horizon-specific granularity $\Delta(H)$
\State Forecast CPU and memory with \modelname{} zero-shot
\State Extract historical and forecast demand summaries
\State Compute effective demand and required capacity via Eq.~\ref{eq:req}
\State Restrict candidate SKUs to the neighbourhood of $s_{i,t}$ defined by $\delta$
\State Retrieve candidate capacities and pricing metadata
\State Assemble structured decision context $c_{i,t,H}$
\State Generate recommendation record from $c_{i,t,H}$
\State Verify grounding of identifiers and numeric outputs
\State \Return $r_{i,t,H}$
\end{algorithmic}
\end{algorithm}

%% file: paper/4-experimental_setup.tex
\section{Experimental Setup}
\label{sec:experiments}

\subsection{Dataset}
We evaluate on production telemetry collected from a dataset of seven VMs, $R_i \in \{R_1, \dots, R_7\}$.
During preprocessing, duplicate entries are removed and missing values are filled by linear interpolation. 
All values are clipped to a minimum of 0.1 to avoid degenerate near-zero inputs that would otherwise destabilise percentage-based error metrics. 
Each metric component is then scaled independently to the $[0,1]$ range using a Min-Max scaler fitted exclusively on the training split and applied unchanged to the validation and test splits.
We study three planning horizons corresponding to the operational recommendation settings considered in this paper: day-scale, week-scale, and month-scale sizing decisions. 
Each horizon is associated with a target resampling frequency and forecast length. 
Specifically, the short-horizon setting uses 30-minute intervals, the medium-horizon setting uses 6-hour intervals, and the long-horizon setting uses 12-hour intervals. 
At each evaluation origin, the model produces a forecast over the relevant horizon, which is then converted into a sizing recommendation using the policy described in Section~\ref{sec:problem}. 
This design allows the same decision pipeline to be evaluated under distinct operational cadences, ranging from near-term review to longer-horizon planning.
For each VM, the time series is partitioned chronologically into training (60\%), validation (20\%), and test (20\%) segments. 
Table~\ref{tab:dataset_stats_raw} summarises the dataset at 1-minute resolution prior to horizon-specific reshaping. 

\begin{table}[htbp]
\centering\caption{Dataset characteristics at 1-min resolution. $N$ indicates the total number of points in the time series.}
\label{tab:dataset_stats_raw}
\renewcommand{\arraystretch}{1.15} % a bit more vertical breathing room
\begin{tabularx}{\columnwidth}{lcccccc}
\toprule
VM & Start & Days & $N$ & \shortstack{CPU avg\\(\%)} & \shortstack{Mem\\(MB)} & \shortstack{Miss. \\(\%)} \\
\midrule
R1 & 2025-03-31 & 196 & 283663 & 35.53 & 6934.97 & 2.70 \\
R2 & 2025-03-31 & 196 & 283678 & 8.82 & 6801.22 & 2.90 \\
R3 & 2025-03-31 & 196 & 283680 & 7.07 & 8338.24 & 0.20 \\
R4 & 2025-03-31 & 196 & 283678 & 3.58 & 13968.74 & 0.70 \\
R5 & 2025-03-31 & 196 & 283680 & 5.65 & 10577.12 & 0.30 \\
R6 & 2025-03-31 & 196 & 283679 & 2.27 & 13814.72 & 1.00 \\
R7 & 2025-03-31 & 196 & 283680 & 10.14 & 15273.05 & 0.90 \\
\bottomrule
\end{tabularx}
\end{table}

\subsection{Forecasting Methods}

We consider a set of classical statistical, decomposition-based, neural, and pre-trained forecasting methods: \textit{Naive Seasonal}, \textit{ARIMA}~\cite{box2015time}, \textit{Prophet}~\cite{taylor2018forecasting}, \textit{TSMixer}~\cite{chen2023tsmixer}, \textit{Chronos-2}~\cite{ansari2024chronos}, and \textit{TimesFM 2.5} (the third checkpoint of TimesFM~\cite{das2024decoder}).
Classical statistical models are fitted on the concatenation of the training and validation splits so that their fitted state aligns with the start of the test window, while for \textit{TSMixer} the validation split is used for model selection and early stopping.
A separate model instance is trained or fitted for each VM--horizon combination. For univariate methods, an independent model is fitted for each metric component.

\noindent\textbf{Naive Seasonal.}
Forecasts are generated by repeating the most recently observed seasonal cycle, with the seasonal period matched to the resampling frequency (48 steps for 30m data, 4 steps for 6h data, and 2 steps for 12h data). At evaluation time, a single deterministic forecast covering the full test interval is produced.

\noindent\textbf{ARIMA}
For each VM, horizon, and metric component, an ARIMA model is selected using \texttt{autoARIMA} from the \texttt{darts} library, which automatically determines the model orders by minimising the Akaike Information Criterion (AIC)~\cite{akaike2003new}. This includes both non-seasonal $(p, d, q)$ and seasonal $(P, D, Q, m)$ components under the library's default configuration. Models are fitted using only the most recent 200 observations. Evaluation is performed using the \texttt{backtest} procedure from \texttt{darts}, which computes historical direct multi-step forecasts over the full target horizon with forecast origins advanced by one step (\texttt{stride}=1).

\noindent\textbf{Prophet.}
Models are fitted independently for each metric component using its default inference procedure (L-BFGS via Stan). Automatic seasonality detection is disabled, and seasonality is specified explicitly according to the data frequency: daily seasonality (periods of 48, 4, and 2 steps for the three settings), together with weekly and monthly components parameterised via low-order Fourier terms. At evaluation time, a single forecast covering the full test interval is produced.

\noindent\textbf{TSMixer.}
Separate models are trained for each (VM, horizon) combination using default architectural and optimisation settings. The input context (lookback window) is set to cover up to one week of history (336, 28, and 14 steps for the three resolutions), and forecasts are generated autoregressively over the target horizon (48, 4, and 2 steps). At evaluation time, rolling historical forecasts are produced from the start of the test split using a stride equal to the forecast horizon, then concatenated and trimmed to align with the test window.

\noindent\textbf{Foundation Models: Chronos-2 and TimesFM 2.5.}
These models are used in a pre-trained setting and do not require additional training or hyperparameter tuning. The input context is set to the maximum available history, or to the maximum context length supported by the model when the history is longer. Evaluation follows the same \texttt{backtest} procedure as ARIMA.

\subsection{Recommendation Layer Evaluation}

In addition to evaluating the forecasting layer, we evaluate the recommendation layer under two zero-shot LLM configurations using the same structured decision prompts. A higher-capacity model is used offline as a \emph{reference model}, generating structured recommendation records from the bounded decision context defined in Section~\ref{sec:system}. A smaller \emph{production model}, chosen to reflect practical latency and cost constraints, is then evaluated on the same prompts in the interactive serving setting. The production model is not trained, fine-tuned, or distilled to imitate the reference model. Instead, the reference model provides a stronger benchmark for recommendation quality, and the resulting \emph{reference agreement} measures how closely a deployable zero-shot model matches stronger-model recommendation behaviour under identical evidence and heuristic constraints.

\subsection{Evaluation Metrics} 
We compute simulated cost by applying \cloudprovider{} on-demand pricing tables to the recommended SKU $\tilde{s}_{i,t,H}$ and comparing this against the cost of the current SKU $s_{i,t}$. 
Positive savings indicate a downgrade recommendation, while negative savings indicate an upgrade. 
This simulation assumes stable on-demand pricing and does not account for migration overhead.
To assess safety in the absence of labelled degradation events, we use an \emph{ex-post exceedance rate}. 
Given a recommendation made at time $t$, we apply the recommended SKU to the realised future telemetry and measure how often the recommended capacity would have been insufficient. 
Formally, let
\[
\mathcal{E}_{i,t+\tau} = \Big[\frac{x^{cpu}_{i,t+\tau}}{100} > \tau^{cpu}\Big] \vee \Big[x^{mem\_used}_{i,t+\tau} > \tau^{mem}\, c^{mem}(\tilde{s}_{i,t,H})\Big].
\]
Then
\begin{equation}
\text{Exceed}_{i,t,H} = \frac{1}{K(H)} \sum_{\tau=1}^{K(H)} \mathbb{I}\left[\mathcal{E}_{i,t+\tau}\right].
\end{equation}

The thresholds $\tau^{cpu}$ and $\tau^{mem}$ are identical to those used in the sizing policy in Section~\ref{sec:problem}, ensuring consistency between decision generation and ex-post evaluation. 
We report mean exceedance across VM--horizon instances together with worst-decile exceedance, since tail behaviour is more operationally consequential than average behaviour. 
For interpretability, exceedance is also reported in minutes per week.
In addition to these decision-level outcomes, we report forecast accuracy as a \textit{secondary diagnostic} using Mean Absolute Percentage Error (MAPE), Mean Absolute Scaled Error (MASE), and Root Mean Squared Error (RMSE). These metrics help characterise model behaviour, but they are not the primary endpoint because the practical objective is safe and cost-effective right-sizing rather than minimising forecast error in isolation.

%% file: paper/5-results.tex
\section{Results}
\label{sec:results}

We evaluate the system first in terms of the provisioning recommendations it induces, and only secondarily in terms of raw forecasting accuracy. This ordering reflects the operational objective of instance sizing: a forecast is useful primarily insofar as it preserves a safe and cost-effective recommendation, rather than merely minimising pointwise error.

\subsection{Recommendation Quality and Operational Impact}

Table II reports simulated monthly costs for the seven VMs in the study. Six of the seven receive a downgrade recommendation, while one is retained at its current SKU. Using the values in Table II, the current configurations total approximately \$1,503/month and the recommended configurations approximately \$708/month, for a combined simulated saving of \$795/month, corresponding to a reduction of about 52.9\%. Savings among downgraded VMs range from \$41 to \$194 per VM per month. These results suggest that the policy identifies substantial over-provisioning across most of the evaluated VMs without defaulting to uniform downsizing: one VM is retained unchanged, while the remaining cases show heterogeneous but consistently positive savings.

\begin{table}[ht]
\centering
\footnotesize
\setlength{\tabcolsep}{3pt}
\caption{Cost analysis of Opus~4.6 reference recommendations. Modal rec.\ = most frequent size across horizons/models/time-points ($N=24$/VM). Costs at 730~hr/mo. $^\dagger$Azure retail West Europe Feb~2026.}
\label{tab:cost_analysis}
\begin{tabular}{@{}llrrrrr@{}}
\toprule
VM & Current & \$/hr & \$/mo & Rec. & \$/mo & Save \\
\midrule
R1 & D4s\_v3$^\dagger$ & .220 & 161 & D4ps\_v5 & 120 & +41 \\
R2 & DS3\_v2$^\dagger$ & .250 & 182 & E2ps\_v5 & 78 & +104 \\
R3 & D4s\_v3$^\dagger$ & .220 & 161 & E2ps\_v5 & 78 & +82 \\
R4 & D4s\_v3$^\dagger$ & .220 & 161 & D2ps\_v5 & 60 & +101 \\
R5 & DS12\_v2$^\dagger$ & .449 & 328 & E4ps\_v5 & 156 & +172 \\
R6 & D4s\_v3$^\dagger$ & .220 & 161 & D2ps\_v5 & 60 & +101 \\
R7 & D8ds\_v5 & .480 & 350 & E4ps\_v5 & 156 & +194 \\
\midrule
\multicolumn{6}{r}{\textbf{Total savings (\$/mo)}} & \textbf{795} \\
\bottomrule
\end{tabular}
\end{table}

\subsection{Cost vs Safety Trade-off}

Table~\ref{tab:exceedance} reports the ex-post exceedance analysis for the five 
downgrade recommendations where exceedance was non-zero or non-trivial. 
Exceedance rates are low across all VMs: three of the five report rates below 
0.2\%, and the highest observed rate is 1.5\% for \texttt{R7}. CPU overflow among exceeding observations is moderate on average but reaches 85.6 percentage points at the 95th percentile for \texttt{R6}, indicating that exceedance, while rare, can be severe when it occurs. Memory overflow is confined to VMs where the recommended SKU reduces available RAM (16\,GB\,$\to$\,8\,GB); in these cases mean overflow ranges from 2.3 to 5.6\,GB. Taken together, the 
results suggest that the sizing policy achieves meaningful cost reductions while 
keeping capacity exceedance rare in aggregate, though the tail behaviour of 
individual VMs warrants attention before acting on downgrade recommendations 
without further review.

\begin{table}[t]
\centering
\scriptsize
\renewcommand{\arraystretch}{1.08}
\setlength{\tabcolsep}{4pt}
\caption{Ex-post exceedance for downgrade recommendations (Opus~4.6
reference) over ${\sim}280$k 1-min observations per VM. \emph{Exc.~(\%)}
= fraction of observations exceeding recommended capacity. Overflow
(Mean, p95) reported over exceeding observations only: CPU in pp above
full utilisation, memory in GB above RAM capacity. ``---'' = zero
exceedance.}
\label{tab:exceedance}
\begin{tabular}{@{}>{\raggedright\arraybackslash}p{3.55cm}rcccc@{}}
\toprule
& & \multicolumn{2}{c}{\textbf{CPU OF}} & \multicolumn{2}{c}{\textbf{Mem OF}} \\
\cmidrule(lr){3-4}\cmidrule(lr){5-6}
\textbf{VM / Rec.\ / Config} & \makecell[c]{\textbf{Exc.}\\\textbf{(\%)}} & \textbf{Mean} & \textbf{p95} & \textbf{Mean} & \textbf{p95} \\
\midrule
\makecell[l]{\texttt{R2} / E2ps\_v5\\{\footnotesize 4$\to$2\,vCPU, 14$\to$16\,GB}} 
& 0.2  & 11.1\,pp & 32.3\,pp & ---     & ---     \\

\makecell[l]{\texttt{R3} / E2ps\_v5\\{\footnotesize 4$\to$2\,vCPU, 16/16\,GB}} 
& 0.5  & 9.0\,pp  & 38.3\,pp & ---     & ---     \\

\makecell[l]{\texttt{R4} / D2ps\_v5\\{\footnotesize 4$\to$2\,vCPU, 16$\to$8\,GB}} 
& $<\!0.1$ & ---      & ---      & 2.3\,GB & 4.2\,GB \\

\makecell[l]{\texttt{R6} / D2ps\_v5\\{\footnotesize 4$\to$2\,vCPU, 16$\to$8\,GB}} 
& $<\!0.1$ & 17.9\,pp & 85.6\,pp & 3.6\,GB & 7.7\,GB \\

\makecell[l]{\texttt{R7} / E4ps\_v5\\{\footnotesize 8$\to$4\,vCPU, 32/32\,GB}} 
& 1.5  & 16.1\,pp & 42.5\,pp & ---     & ---     \\
\bottomrule
\end{tabular}
\end{table}

\subsection{Forecast Accuracy Results}
Tables~\ref{tab:forecast-mem} and~\ref{tab:forecast-cpu} present the complete
forecasting results for all evaluated models under rolling-origin backtesting.
We report mean $\pm$ standard deviation across VMs for MAPE, MASE, and RMSE
at the Day, Week, and Month horizons. These results complement the main text
by providing a full comparison of statistical, supervised, and foundation-based
approaches.

Despite no task-specific retraining, \textit{Chronos-2} and \textit{TimesFM 2.5} achieve consistently
competitive performance across all horizons. While they do not systematically
minimize all error metrics, their accuracy remains comparable to supervised
approaches. Given that forecasting serves as a diagnostic component within
CLOUDADV, these results suggest that zero-shot models provide sufficiently
reliable signals to support downstream sizing recommendations without per-VM
retraining.

\begin{table}[t]
\caption{Forecasting accuracy for \textbf{Available Memory} (mean\,$\pm$\,std
across VMs) under rolling-origin backtesting. Step size: Day\,=\,30\,min,
Week\,=\,6\,h, Month\,=\,12\,h. Best per horizon/metric in \textbf{bold}.}
\label{tab:forecast-mem}
\centering
\footnotesize
\setlength{\tabcolsep}{4pt}
\begin{tabular}{@{} ll rrr @{}}
\toprule
Horizon & Model & MAPE\,(\%) & MASE & RMSE\,(MB) \\
\midrule
\multirow{6}{*}{\textit{Day}}
& ARIMA       & 6.01\,$\pm$\,9.17  & \textbf{5.07\,$\pm$\,4.04} & 443.98\,$\pm$\,467.65  \\
& Chronos-2   & 4.98\,$\pm$\,9.32 & 5.76\,$\pm$\,8.42 & 325.88\,$\pm$\,397.55 \\
& Naive Seas. & 15.76\,$\pm$\,29.71 & 20.62\,$\pm$\,29.89 & 1126.11\,$\pm$\,1348.18 \\
& Prophet     & 21.98\,$\pm$\,39.87 & 29.58\,$\pm$\,45.21 & 1424.20\,$\pm$\,1713.14 \\
& TSMixer     & 8.79\,$\pm$\,17.40  & 13.00\,$\pm$\,22.87 & 733.57\,$\pm$\,1094.59  \\
& TimesFM 2.5 & \textbf{4.90 $\pm$ 9.16} & 5.61 $\pm$ 8.33 & \textbf{321.91 $\pm$ 394.06} \\
\midrule
\multirow{6}{*}{\textit{Week}}
& ARIMA       & 12.05\,$\pm$\,23.83 & \textbf{3.62\,$\pm$\,1.65} & 801.57\,$\pm$\,1139.38 \\
& Chronos-2   & 11.62\,$\pm$\,22.31 & 4.49\,$\pm$\,4.41 & 792.31\,$\pm$\,1069.63 \\
& Naive Seas. & 15.34\,$\pm$\,29.05 & 6.75\,$\pm$\,6.57 & 1086.71\,$\pm$\,1329.87 \\
& Prophet     & 16.96\,$\pm$\,30.24 & 8.15\,$\pm$\,6.87 & 1176.59\,$\pm$\,1298.69 \\
& TSMixer     & \textbf{8.84\,$\pm$\,17.13} & 3.79\,$\pm$\,4.66 & \textbf{685.13\,$\pm$\,907.11} \\
& TimesFM 2.5 & 11.11 $\pm$ 22.00 & 4.19 $\pm$ 4.26 & 741.36 $\pm$ 1044.24 \\
\midrule
\multirow{6}{*}{\textit{Month}}
& ARIMA       & 11.33\,$\pm$\,19.71 & 3.46\,$\pm$\,2.57 & 888.58\,$\pm$\,1107.68 \\
& Chronos-2   & 11.02\,$\pm$\,19.89 & 3.28\,$\pm$\,2.65 & 835.67\,$\pm$\,1046.34 \\
& Naive Seas. & 13.93\,$\pm$\,25.60 & 4.18\,$\pm$\,3.70 & 1016.48\,$\pm$\,1209.41 \\
& Prophet     & 17.50\,$\pm$\,31.02 & 5.59\,$\pm$\,4.92 & 1303.39\,$\pm$\,1439.35 \\
& TSMixer     & \textbf{8.78\,$\pm$\,15.36} & \textbf{2.67\,$\pm$\,2.78} & \textbf{767.35\,$\pm$\,918.13} \\
& TimesFM 2.5 & 11.57 $\pm$ 21.72 & 3.20 $\pm$ 2.43 & 795.30 $\pm$ 990.04 \\
\bottomrule
\end{tabular}
\end{table}

\begin{table}[t]
\caption{Forecasting accuracy for \textbf{Percentage CPU} (mean\,$\pm$\,std
across VMs) under rolling-origin backtesting. Step size: Day\,=\,30\,min,
Week\,=\,6\,h, Month\,=\,12\,h. Best per horizon/metric in \textbf{bold}.}
\label{tab:forecast-cpu}
\centering
\footnotesize
\setlength{\tabcolsep}{4pt}
\begin{tabular}{@{} ll rrr @{}}
\toprule
Horizon & Model & MAPE\,(\%) & MASE & RMSE \\
\midrule
\multirow{6}{*}{\textit{Day}}
& ARIMA       & 29.13\,$\pm$\,20.04 & 1.84\,$\pm$\,0.83 & 6.40\,$\pm$\,8.05 \\
& Chronos-2   & \textbf{12.72\,$\pm$\,7.49} & \textbf{0.92\,$\pm$\,0.19} & \textbf{4.25\,$\pm$\,4.59} \\
& Naive Seas. & 27.87\,$\pm$\,15.12 & 1.75\,$\pm$\,0.39 & 7.27\,$\pm$\,7.37 \\
& Prophet     & 34.24\,$\pm$\,22.69 & 2.01\,$\pm$\,0.75 & 7.23\,$\pm$\,8.67 \\
& TSMixer     & 18.79\,$\pm$\,11.45 & 1.26\,$\pm$\,0.38 & 5.96\,$\pm$\,6.35 \\
& TimesFM 2.5 & 13.82 $\pm$ 8.37 & 0.99 $\pm$ 0.23 & 4.41 $\pm$ 4.85 \\
\midrule
\multirow{6}{*}{\textit{Week}}
& ARIMA       & 19.91\,$\pm$\,15.33 & 1.07\,$\pm$\,0.65 & 4.65\,$\pm$\,7.12 \\
& Chronos-2   & \textbf{12.74\,$\pm$\,9.00} & \textbf{0.63\,$\pm$\,0.28} & \textbf{3.66\,$\pm$\,5.52} \\
& Naive Seas. & 23.53\,$\pm$\,18.39 & 1.03\,$\pm$\,0.41 & 5.83\,$\pm$\,8.98 \\
& Prophet     & 27.65\,$\pm$\,19.36 & 1.34\,$\pm$\,1.05 & 5.31\,$\pm$\,8.03 \\
& TSMixer     & 16.37\,$\pm$\,15.85 & 0.71\,$\pm$\,0.29 & 4.91\,$\pm$\,7.86 \\
& TimesFM 2.5 & 13.13 $\pm$ 9.14 & 0.65 $\pm$ 0.27 & 3.77 $\pm$ 5.70 \\
\midrule
\multirow{6}{*}{\textit{Month}}
& ARIMA       & 22.49\,$\pm$\,15.48 & 0.84\,$\pm$\,0.46 & 4.75\,$\pm$\,7.03 \\
& Chronos-2   & 16.61\,$\pm$\,8.51  & 0.58\,$\pm$\,0.25 & 4.15\,$\pm$\,5.95 \\
& Naive Seas. & 20.91\,$\pm$\,12.54 & 0.68\,$\pm$\,0.31 & 4.96\,$\pm$\,6.88 \\
& Prophet     & 21.17\,$\pm$\,8.25  & 0.80\,$\pm$\,0.63 & 4.26\,$\pm$\,5.62 \\
& TSMixer     & \textbf{15.01\,$\pm$\,7.08} & \textbf{0.56\,$\pm$\,0.30} & 4.43\,$\pm$\,6.24 \\
& TimesFM 2.5 & 17.41 $\pm$ 9.58 & 0.59 $\pm$ 0.24 & \textbf{4.15 $\pm$ 5.85} \\
\bottomrule
\end{tabular}
\end{table}

\subsection{Reference--Production Recommendation Alignment}
\label{sec:llm_consistency}

To assess whether a deployable zero-shot model remains aligned with stronger-model recommendation behaviour, we compare recommendations produced by the higher-capacity reference model and the smaller production model under identical structured decision prompts. Table~\ref{tab:action_distribution} shows the resulting action distribution by forecasting model and planning horizon. Across all horizon--model conditions, \texttt{downgrade} recommendations dominate, accounting for 75--86\% of decisions, while \texttt{maintain} remains fixed at 14\% and \texttt{upgrade} is uncommon. Chronos-2 and TSMixer induce closely similar action distributions at each horizon despite differences in raw forecasting accuracy, which further supports the paper’s decision-aligned framing.

Table~\ref{tab:model_accuracy} reports agreement between the reference and production models. Agreement remains high overall, but the updated table supports a more precise characterization: exact recommended-size agreement ranges from 78.6\% to 100.0\%, while action-level agreement ranges from 85.7\% to 100.0\%. Agreement is strongest at the day and week horizons, and the main degradation appears at the month horizon, where exact-size agreement falls to 78.6\% for both forecasting models and action-level agreement falls to 89.3\% for TSMixer and 85.7\% for Chronos-2.

This pattern is consistent with the broader forecasting results: longer-horizon decisions are more sensitive to modest shifts in forecast summaries, especially when the bounded decision context lies near the boundary between maintaining and resizing. In such cases, small differences in estimated demand or in model interpretation of the structured context can change the recommended target SKU even when the overall action tendency remains similar. The month horizon is therefore the most informative stress case for deployment alignment rather than evidence of a systematic breakdown.

Taken together, the results suggest that the smaller production model remains closely aligned with the stronger reference model in the present setting, especially for confidently overprovisioned VMs. Although supervised training improves some longer-horizon forecasting metrics, those gains do not materially alter the resulting recommendation distribution in most cases. From a deployment perspective, this supports zero-shot forecasting together with lower-cost recommendation serving as a practical option in environments where per-VM retraining and high-capacity online inference are costly or impractical.

\begin{table}[ht]
\centering
\caption{Reference recommendation action distribution (Opus~4.6, $n=168$)
by horizon and forecasting model. Percentages within each horizon--model group.}
\label{tab:action_distribution}
\begin{tabular}{llrrr}
\toprule
Horizon & Forecast Model & Downgrade & Maintain & Upgrade \\
\midrule
  \textbf{Day} & TSMixer & 21 (75\%) & 4 (14\%) & 3 (11\%) \\
   & Chronos2 & 22 (79\%) & 4 (14\%) & 2 (7\%) \\
\addlinespace
  \textbf{Week} & TSMixer & 22 (79\%) & 4 (14\%) & 2 (7\%) \\
   & Chronos2 & 22 (79\%) & 4 (14\%) & 2 (7\%) \\
\addlinespace
  \textbf{Month} & TSMixer & 24 (86\%) & 4 (14\%) & 0 (0\%) \\
   & Chronos2 & 23 (82\%) & 4 (14\%) & 1 (4\%) \\
\bottomrule
\end{tabular}
\end{table}

\begin{table}[ht]
\centering
\caption{Reference--production recommendation agreement (Qwen3.5-35b
vs.\ Opus~4.6) by horizon and forecasting model. \emph{Size} = exact
match on recommended VM size; \emph{Action} = match on action label
(downgrade/upgrade/maintain).}
\label{tab:model_accuracy}
\setlength{\tabcolsep}{5pt}
\begin{tabular}{llrrr}
\toprule
Horizon & Model & $N$ & Size Agr.\ (\%) & Action Agr.\ (\%) \\
\midrule
Day & TSMixer & 28 & 100.0 & 100.0 \\
Day & Chronos2 & 28 & 92.9 & 96.4 \\
\addlinespace
Week & TSMixer & 28 & 96.4 & 96.4 \\
Week & Chronos2 & 28 & 96.4 & 96.4 \\
\addlinespace
Month & TSMixer & 28 & 78.6 & 89.3 \\
Month & Chronos2 & 28 & 78.6 & 85.7 \\
\bottomrule
\end{tabular}
\end{table}

\subsection{Engineer Audit of Recommendation Faithfulness}

We performed a stratified engineer audit of 17 recommendations (${\approx}10\%$ of $n{=}168$) to assess the faithfulness of the reference recommendation layer. For each audited case, an engineer re-derived a baseline recommendation from the candidate list by selecting the lowest-cost VM under the stated CPU and memory requirements and sizing rules.

The audit yielded a \textbf{17/17} overall pass rate. In \textbf{15} cases, the LLM recommendation exactly matched the engineer-derived baseline. In the remaining \textbf{2} cases, the selected VM differed from the baseline but was still judged acceptable by the engineer. This suggests that, within the bounded recommendation setting used here, the reference model generally follows the intended sizing policy and candidate constraints, with only minor variation in borderline cases.

%% file: paper/7-conclusion.tex
\section{Discussion and Conclusion}
\label{sec:conclusion}

We presented \systemname{}, an engineer-facing advisory system for cloud instance sizing under workload drift, and evaluated whether zero-shot foundation model forecasts can support safe and cost-effective provisioning within a bounded workflow.
Our results suggest that they can. 
Chronos-2 achieves the strongest short- and medium-horizon CPU performance, remains competitive for memory forecasting, and produces
recommendation patterns close to those of a supervised per-VM baseline. At the decision level,
\systemname{} reduces estimated monthly spend from \$1,503 to \$708 across the evaluated VM set (52.9\%), with downgraded cases peaking at 1.5\% ex-post exceedance.

Agreement between the smaller production model and the stronger reference model is high,
with exact SKU agreement ranging from 78.6\% to 100.0\% and action-level agreement from 85.7\% to 100.0\%.
Longer horizons, especially the month horizon, expose the main stress case: recommendations become more sensitive to modest forecast shifts near maintain-versus-resize boundaries.
The least reliable cases are those involving reduced RAM.

Memory exceedance is concentrated in recommendations that reduce available RAM, where abrupt application-level changes, such as cache growth, additional services, or traffic shifts, can cause substantial tail overflow.
CPU exceedance is less frequent, but,
tends to appear as occasional large misses rather than persistent mild underprovisioning.
More broadly, \systemname{} remains vulnerable to non-stationarity outside the historical context window, including migrations, long business cycles, and application changes.
These failure modes reinforce its intended role as
an advisory mechanism requiring engineer review, particularly for memory-sensitive workloads, recently changed systems, and recommendations near resizing boundaries.

Several limitations of the present study should be noted. The evaluation covers seven production VMs, which constrains the assessment of generalizability to more diverse workload profiles and VM families. Workload drift is addressed conceptually rather than through controlled experimental manipulation, so the claimed advantages of zero-shot models in non-stationary settings remain to be validated at scale. The pipeline also lacks a direct ablation against a fully rule-based system, which would more precisely quantify the contribution of the LLM layer beyond the sizing policy alone; we leave this comparison to future work.
In practice, \systemname{} is best suited to
recommendations alongside knowledge
of upcoming workload changes.
Short horizons help validate near-term safety, while week- and month-scale horizons support recurring capacity reviews and purchasing decisions.
The cost analysis highlights a common production pattern: conservative SKUs often persist despite heterogeneous resource demands, and better-matched newer families can yield meaningful aggregate savings. 
For teams unable to maintain per-VM supervised forecasting pipelines, zero-shot forecasting can still provide useful decision support.
Reference--production agreement further suggests that the user-facing layer can use a smaller deployment model while preserving much of the stronger model's behaviour.
This positions \systemname{} as a complement to existing cost-visibility tooling, adding forward-looking, forecast-grounded recommendations to retrospective utilisation summaries.
Future work will extend evaluation to more heterogeneous VM populations, incorporate richer contextual signals to improve robustness under workload shift, and explore tighter integration with continual adaptation mechanisms for settings where periodic zero-shot re-evaluation is insufficient.